\pdfoutput=1

\documentclass[11pt]{article}

\usepackage{acl}
\usepackage{float}

\usepackage{times}
\usepackage{latexsym}
\usepackage{multirow}
\usepackage{subfigure}
\usepackage{booktabs}
\usepackage[T1]{fontenc}
\usepackage[utf8]{inputenc}
\usepackage{microtype}
\usepackage{todonotes}
\usepackage{array}

\usepackage[ruled, linesnumbered]{algorithm2e}
\SetAlFnt{\small}
\SetAlCapFnt{\small}
\SetAlCapNameFnt{\small}

\title{Curriculum Learning for Graph Neural Networks: A Multiview Competence-based Approach}

\author{Nidhi Vakil \\
  Department of Computer Science \\
  University of Massachusetts Lowell \\
  \texttt{nvakil@cs.uml.edu} \\ \And
  Hadi Amiri \\
  Department of Computer Science \\
  University of Massachusetts Lowell \\
  \texttt{hadi@cs.uml.edu} \\}

\begin{document}
\maketitle
\begin{abstract}
A curriculum is a planned sequence of learning materials and an effective one can make learning efficient and effective for both humans and machines. Recent studies developed effective data-driven curriculum learning approaches for training graph neural networks in language applications. However, existing curriculum learning approaches often employ a single criterion of difficulty in their training paradigms. In this paper, we propose a new perspective on curriculum learning by introducing a novel approach that builds on graph complexity formalisms (as difficulty criteria) and model competence during training. The model consists of a scheduling scheme which derives effective curricula by accounting for different views of sample difficulty and model competence during training. The proposed solution advances existing research in curriculum learning for graph neural networks with the ability to incorporate a fine-grained spectrum of graph difficulty criteria in their training paradigms. Experimental results on real-world link prediction and node classification tasks illustrate the effectiveness of the proposed approach.\footnote{Code, data splits and guidelines are available at \url{https://clu.cs.uml.edu/tools.html}.}

\end{abstract}

\section{Introduction}

Graph Neural Networks (GNNs) are generally trained using stochastic gradient descent (SGD), where the standard approach is to iteratively use the {\em entire} training data to optimize model's objective until convergence. Curriculum learning techniques improve this training process by scheduling examples for training, e.g., by gradually learning from easier examples before training with harder ones. Such curricula can be predefined by humans~\cite{bengio2007scaling,bengio2009curriculum} or dynamically derived from data during training~\cite{jiang2018mentornet,castells2020superloss}.

Curriculum learning for graph neural networks is an emerging area of research. Recently, \citet{chu2021cuco} employed a traditional curriculum learning approach introduced in~\cite{bengio2009curriculum} to improve negative sampling for graph classification. \citet{wang2021curgraph} proposed to estimate the difficulty of graph entities--nodes, edges or subgraphs--based on the intra- and inter-class distributions of their embeddings in supervised settings, and developed a smooth-step function to gradually introduce harder examples to GNNs during training. \citet{nidhi-etal-2022-gtnn} developed a loss-based curriculum learning approach that dynamically adjusts the difficulty boundaries of training samples based on their sample-level loss trajectories obtained from recent training dynamics of GNN models. 

\begin{table*}[!t]\small
    \centering
    \begin{tabular}{c c p{7cm} c c  c }
     & & & & & \\
         \textbf{ID} & \textbf{subgraph} & \textbf{Sentence} & \textbf{Label} &  \textbf{Degree} & \textbf{Centrality} \\
         \toprule
         G1 & \raisebox{-.9\height}{\includegraphics[width = 2.3cm, height = 2.3cm]{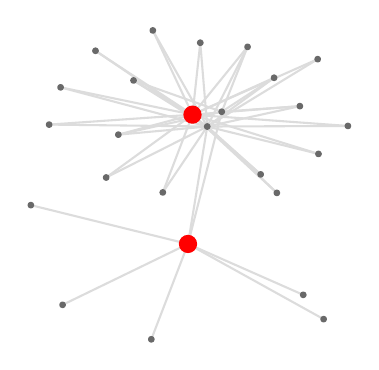}}& Prominent examples include mutations in the transporters for dopamine (DAT, \underline{\bf SLC6A3}), for creatine (CT1, SLC6A8), and for glycine (GlyT2, SLC6A5), which result in infantile dystonia, mental retardation, and \underline{\bf hyperekplexia}, respectively. & False &  24 & 1.14\\
           G2 & \raisebox{-.7\height}{\includegraphics[width = 2.3cm, height = 2.3cm]{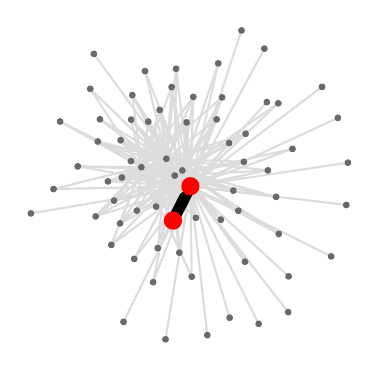}\label{fig:pgr_2}} &Depletion of \underline{\bf PRODH} and GSALDH in humans leads to \underline{\bf hyperprolinemia}, which is associated with mental disorders such as schizophrenia. & True &  69 & 1.46 \\

           G3 & \raisebox{-.8\height}{\includegraphics[width = 2.3cm, height = 2.3cm]{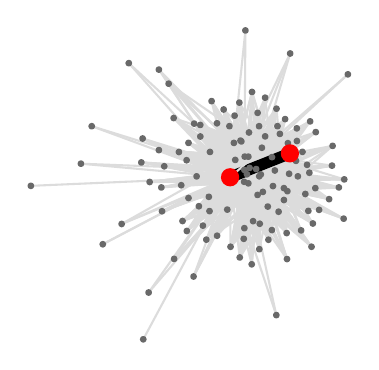}\label{fig:pgr_3}} & Lately, \underline{\textbf{ARMC5}} was linked to the cyclic AMP signaling pathway, which could be implicated in all of mechanisms of cortisol-secreting by macronodules \underline{\textbf{adrenal hyperplasia}} and the molecular defects in: G protein aberrant receptors; MC2R; GNAS; PRKAR1A; PDE11A; PDE8B. & True &  122 & 1.40\\
    \bottomrule
    \end{tabular}
    \caption{The difficulty of training examples--target node pairs in red--can be assessed based on their subgraphs, k-hop neighbors of target nodes. For brevity, we show two views: node degree and closeness centrality. Boldfaced tokens indicate nodes in the graph, Label indicates if the sentence reports a causal relation between the nodes, and Degree and Centrality report the sum of degree and closeness centrality scores of the target nodes in their subgraphs. Each subgraph provides a structural view of the target nodes in sentences. The relative difficulty of examples is different across views, e.g., G2 is less difficult than G3 according to Degree but more difficult according to Centrality.}
    \label{tab:pgr_example}
\end{table*}

To the best of our knowledge, existing curriculum learning approaches often employ a {\em single} criterion of difficulty in their curriculum learning framework, e.g., 
prediction loss~\citep{wu2021when}, 
consistency in prediction loss~\citep{xu-etal-2020-curriculum}, 
moving average of loss~\citep{zhou2020curriculum} or 
transformations of loss~\citep{nidhi-etal-2022-gtnn}. 
We address this gap by developing a new curriculum learning approach for GNNs titled \textbf{M}ultiview \textbf{C}ompetence-based \textbf{C}urriculum \textbf{L}earning (MCCL) that builds on the complexity formalisms of graph data. By leveraging rich graph structures, graph complexity formalisms and model \textit{competence} (learning progress), we will design robust curricula for training GNNs. Table~\ref{tab:pgr_example} shows three subgraphs ranked differently according to different graph complexity indices. If complexity is measured by \textit{node degree}, then G1 and G2 are less complex than G3 because target nodes in these subgraphs have an overall smaller node degrees. However, if complexity is measured by \textit{closeness centrality}\footnote{Closeness centrality~\cite{sabidussi1966centrality} is smaller for central nodes--those that are closer to other nodes in the graph.}, then G2 is more complex than G3 because the target nodes are less central in G2 than those in G3. It is evident that complexity indices (views) can vary significantly in their difficulty estimates of graph data. 

The objective of this work is improve the training process of GNNs by strategically and dynamically (during training) prioritizing key complexity indices, aiming to guide the model toward better minima within its parameter space.
Graph complexity is a well-established area of research and our review of relevant literature suggests that there exist various techniques that employ structural properties of nodes, edges and subgraphs to quantify complexity of graph data~\cite{KIM20082637,vishwanathan2010graph,newman2018networks,kriege2020survey}. We build on these indices to design our curriculum learning framework which treats each complexity index as a view of difficulty. Our approach consists of a novel data scheduling scheme which derives effective curricula based on given views of sample difficulty and model competence during training. Specifically, given a downstream GNN model, our data scheduler gradually selects training examples from a graph complexity view based on competency of the GNN model during training. The model updates its competency and the scheduler determines the next best view for training the model. As model competency gradually increases, the scheduler allows using more signals from different views.  






The contributions of this paper are as follows:
\begin{itemize}
    \item A new curriculum learning approach that effectively leverages complexity formalisms of graph data, taking into account multiview difficulty of training data samples and model's learning progress, and
    \item Key insights into important complexity indices for effective training of graph neural networks for NLP applications. 
\end{itemize}

We conduct extensive experiments on real world datasets for link prediction and node classification tasks in text graph datasets. Our approach results in 3.3 and 1.8 absolute points improvements in F1-score over the state-of-the-art model on link prediction datasets and 6.7 and 4.9 absolute points improvement on the node classification dataset. 
The results show that the contribution of complexity indices in training depends on factors such as training stage and model behavior. When the scheduling criterion relies solely on complexity indices, the scheduler tends to initially focus on indices that operate locally around nodes, and later shifts to those that operate globally at graph level. Extending schedulers based on model dynamics (e.g., loss) results in both local and global indices being used throughout the training. 
These findings provide insights into the type of complexity information that GNNs learn at different stages of their training.

    

\section{Competence-based Multiview Curricula}
\label{sec:method}
We present a competence-based multiview curriculum learning framework for training GNNs. At every training iteration, the framework selects a sub-set of training examples based on the best complexity index (view) and model's competence at that iteration. Algorithm~\ref{algo:ccl_multi_view} describes the overall approach. We first introduce our complexity indices and then present the model.

\subsection{Graph Complexity Formalisms} \label{sec:metric_def}
Various graph complexity indices were introduced in graph theory~\citep{kashima2003marginalized,borgwardt2005shortest,vishwanathan2010graph,kriege2020survey,newman2018networks}. 
We consider 26 of such indices which represent criteria of difficulty in our curriculum learning framework.\footnote{Our list of complexity indices may not be exhaustive. However, our framework and implementation allows adding any number of additional indices.} In what follows, we describe a few representative complexity indices and refer the reader to Appendix~\ref{sec:appendix} for a full description of all indices. 

Since GNNs train through neural message passing at subgraph level~\cite{gilmer2017neural,hamilton2017inductive}, we compute complexity indices with respect to the $k$-hop neighbors (subgraph) of target nodes. For tasks involving two nodes (e.g., relation extraction), we sum the scores computed for the node pairs. We use Networkx~\cite{hagberg2008exploring} to compute the indices:


\begin{itemize}
\itemsep0em 
    \item \textbf{Degree:} The number of immediate neighbors of a node in a graph. 
    \item \textbf{Average neighbor degree:} Average degree of the neighbors of a node: 
    \begin{eqnarray*}
    \frac{1}{|\mathcal{N}_i|}\sum_{j\in \mathcal{N}_i} k_j,
    \end{eqnarray*}
    where $\mathcal{N}_i$ is the set of neighbors of node $i$ and $k_j$ is the degree of node $j$. 
    
    
    \item \textbf{Katz centrality:} The centrality of a node computed based on the centrality of its neighbors. Katz centrality computes the relative influence of a node within a network by measuring the number of immediate neighbors and number of walks between node pairs. It is computed as follows:
    \begin{eqnarray*}
    x_{i} = \alpha\sum_{j}\mathbf{A}_{ij}x_{j}+\beta,
    \end{eqnarray*}
    where $x_i$ is the Katz centrality of node $i$, $\mathbf{A}$ is the adjacency matrix of Graph $G$ with eigenvalues $\lambda$. The parameter $\beta$ controls the initial centrality and $\alpha$ $<$ 1 / $\lambda_{max}$.
    
    
    
    
    
    
    
    \item \textbf{Resource allocation index:} For nodes $i$ and $j$ in a subgraph, the resource allocation index is defined as follows: 
    \begin{eqnarray*}
    \sum_{k \in (\mathcal{N}_i\bigcap\mathcal{N}_j)}\frac{1}{|\mathcal{N}_k|},
    \end{eqnarray*}
    which quantifies the closeness of target nodes based on their shared neighbors.
    
    \item \textbf{Subgraph density:} The density of an undirected subgraph is computed as follows:
    \begin{eqnarray*}
    \frac{e}{v(v-1)},
    \end{eqnarray*}
    where $e$ is the number of edges and $v$ is the number of nodes in the subgraph.
    
    \item \textbf{Local bridge:} A local bridge is an edge that is not part of a triangle in the subgraph. We take the number of local bridges in a subgraph as a complexity index.
    
    
    
    
    
    \item \textbf{Subgraph connectivity:} Is measured by the {\em minimum} number of nodes that must be removed to disconnect the subgraph.
    
    \item \textbf{Eigenvector centrality:} Eigenvector centrality computes the centrality for a node based on the centrality of its neighbors. The eigenvector centrality for node i is $Ax$ = $\lambda x $. where $A$ is the adjacency matrix of the graph $G$ with eigenvalue $\lambda$.
    
    

\end{itemize}

We note that our approach does not depend on any specific index. However, we recommend considering indices that are computationally inexpensive for applicability to large graphs. The complexity scores of each index are normalized into [0, 1] range using L2 norm.

\subsection{Model Competency }
We define model competence at each training iteration $t$ as the fraction of training data that can be used by the model at time $t$; we refer to this fraction by $c(t)$. Our curriculum learning framework employs difficulty indices to select $c(t)$ fraction of examples to train its downstream model (a GNN). 
We employ the following function~\citep{platanios2019competence} to quantify competence:
\begin{eqnarray}
c(t)=\min\left(1,\sqrt[p]{t\left(\frac{1-c_{0}^{p}}{T}\right)+c_{0}^{p}}\right),
\label{eq:competence}
\end{eqnarray}
where $t$ is the training iteration, $p$ controls the sharpness of the curriculum so that more time is spent on the examples added later in the training, $T$ is the maximum curriculum length (number of iterations), and $c_0$ is the initial value of the competence. $c(t)$ gradually increases to achieve the maximum value of 1, which covers the entire training dataset. We set $p= 2$ and $c_0 = 0.01$ as suggested in~\cite{platanios2019competence}.

\subsection{Prioritizing Important Difficulty Indices}\label{sec:prioritizing}

Difficulty indices vary significantly in their difficulty estimates, owing to the complicated topology and indistinct patterns in graph data. Our framework strategically prioritizes key difficulty indices while training a GNN model. Specifically, the framework employs two mechanisms (see line 7 in Algorithm \ref{algo:ccl_multi_view}) to determine which index (i.e., top $c(t)$ portion of training data ranked by the difficulty scores obtained from the index) should be used for training the downstream GNN model at iteration $t$: (i) model-based and (ii) index-based approaches:

\paragraph{Model-based:} This approach performs a forward pass on the selected portion of training data and calculates the average loss of the GNN on these examples. The index with the maximum (or minimum, depending on the curriculum) average loss will be selected at iteration $t$ and it's top $c(t)$ examples will be used for training the downstream GNN. Minimum average loss prioritizes easier examples over harder ones for training. On the other hand, maximum average loss prioritizes harder examples (as in an anti-curriculum setting).

\paragraph{Index-based:} This approach uses the actual difficulty scores obtained from indices. The index with minimum (or maximum) average difficulty score across its top $c(t)$ portion of training samples will be selected for training and calculating the error (see lines 9-14 in Algorithm \ref{algo:ccl_multi_view}). We note that the index-based approach is computationally inexpensive compared to the model-based approach, and results in comparable performance, see results in experiments (Tables~\ref{tab:ab_study}) 

\subsection{Base Graph Neural Network Model}\label{sec:gtnn}
Our approach is model agnostic and can be applied to any GNN. We use the graph-text neural network (GTNN) model\footnote{\url{https://github.com/CLU-UML/gtnn}} from~\cite{nidhi-etal-2022-gtnn} as the base model because it is designed for text-graph data. The model integrates textual information with graph structure and directly uses text embeddings at prediction layer to avoid information loss in the iterative process of training GNNs. We use this model as a base model to compare our and baseline curriculum learning approaches on graph data.

\begin{algorithm}[t]

\SetKwData{Left}{left}\SetKwData{This}{this}\SetKwData{Up}{up}
\SetKwFunction{Union}{Union}\SetKwFunction{FindCompress}{FindCompress}
\SetKwInOut{Input}{input}\SetKwInOut{Output}{output}
\Input{ \\ 
        D: Training data of size $n$ \\ 
        L: Difficulty indices \\
        M: GNN Model \\
        O: easy-to-hard vs. hard-to-easy transition
      }
\Output{ Trained model M$^*$}
\BlankLine

Compute complexity scores for each index $i$ in L and store the results in L$_i$\\
L$_i \leftarrow sort($L$_i)$ \textit{\# in ascending or descending order}. \\ 
\For{$t\leftarrow 0$ \KwTo $T$}{
    $c(t) \leftarrow$ competence from Eq~(\ref{eq:competence})\\
    \ForEach{ index in L}{
        $l_i$ $\leftarrow$ top $(c(t) \times n)$ examples from L$_i$\\
        $e_i\leftarrow$ average loss or complexity of $l_i$ \\
    }
    \eIf {O = easy-to-hard} {$j$ = $\arg\min_i e_i$} {$j$ = $\arg\max_i e_i$}
    Train M with $l_j$ samples
}
\caption{Multiview Competence-based Curriculum Learning (MCCL).}
\label{algo:ccl_multi_view}
\end{algorithm}

\section{Experimental Results}\label{sec:datasets}
\subsection{Datasets}
\paragraph{Gene Phenotype Relation (PGR)} ~\cite{sousa2019silver}: PGR is created from PubMed articles and contains sentences describing causal relations between genes and phenotypes (symptoms); see Table~\ref{tab:pgr_example} for examples of this dataset.

\paragraph{Gene, Disease, Phenotype Relation (GDPR)}~\cite{nidhi-etal-2022-gtnn}: GDPR contains different types of relations among genes, diseases and phenotypes, and long texts describing them.  

\paragraph{Cora}\cite{mccallum2000automating}: Cora is a relatively small citation network, in which nodes are scientific papers and edges are citations among them. Each paper is categorized into one of the seven subject categories and is provided with a textual feature word vector obtained from the content of the paper. 

\paragraph{Ogbn-arxiv}~\cite{hu2020open}: This Open Graph Benchmark dataset is a citation network between papers in the Computer Science domain. Each node in the graph is a paper and an edge represents a citation from one paper to another. Also, each paper contains 128 dimension embedding vector obtained by taking the average of the words present in the title and the abstract.

Table~\ref{tab:data} shows the statistics of the above datasets. We use PGR and GDPR for link prediction and Cora  and Ogbn-Arxiv for node classification. 

\begin{table}[t]\small
    \centering
    \begin{tabular}{l c c c c} 
         &\textbf{ GDPR} &\textbf{PGR} & \textbf{Cora} & \textbf{Ogbn-Arxiv} \\ \toprule
         \textbf{Nodes} & 18.3K & 20.4K & 2.7K & 169K\\
         \textbf{Edges} & 365K & 605K & 5.4K & 1.1M\\ \midrule
    \textbf{Train}     &  30.1K & 2.6K & 2.1K & 90K \\
    \textbf{Test} & 3.7K & 155 & 271 & 49K \\
    \textbf{Val} & 3.7K & -- & 271 & 30K \\ \bottomrule
    \end{tabular}
    \caption{Dataset statistics.}
    \label{tab:data}
\end{table}

\subsection{Baselines} \label{sec:baselines}

\paragraph{CurGraph}~\cite{wang2021curgraph} is a curriculum learning framework for graphs that computes difficulty scores based on the intra- and inter-class distributions of embeddings and develops a smooth-step function to gradually include harder samples in training. We report the results of our implementation of this approach.   
    
\paragraph{SuperLoss} (SL)~\cite{castells2020superloss} is a generic curriculum learning approach that dynamically learns a curriculum from model behavior. It uses a fixed difficulty threshold at batch level, determined by the exponential moving average of all sample losses, and assigns higher weights to easier samples than harder ones.
    
\paragraph{Trend-SL}~\cite{nidhi-etal-2022-gtnn} is a curriculum learning approach which extends~\cite{castells2020superloss} by incorporating sample-level loss trends to better discriminate easier from harder samples and schedule them for training.


\subsection{Settings}

We consider 1-hop neighbors for PGR and GDPR and 2-hop neighbors for Cora and Ogbn-Arxiv to create subgraphs for computing complexity indices, see Section~\ref{sec:metric_def}, and training the GTNN model, see Section~\ref{sec:gtnn}. We train all models for a maximum number of $100$ iterations for PGR and GDPR, and $500$ iterations for Cora and Ogbn-Arxiv with model checkpoint determined by validation data for all models. We conduct all experiments using Ubuntu 18.04 on a single 40GB A100 Nvidia GPU. 

We consider 26 complexity indices listed in Appendix~\ref{sec:appendix}. Since some of the indices are highly co-related, we use k-means to group them based on the Pearson co-relations between their ranking of training samples. We categorize indices into 10 clusters through grid search, which effectively prevents any redundancy in the index space. We randomly select an index from each cluster to be used by our curriculum learning framework. Indices that are used by the framework are labeled by asterisks in Appendix~\ref{sec:appendix}.


We report F1 score (on positive class) for PGR and GDPR datasets, and Accuracy score for Cora and Ogbn-Arxiv datasets. In addition, we
use t-test for significance testing and asterisk mark (*) to indicate significant difference at $\rho = 0.01$.

\begin{table}\small
  \centering
  
    \begin{tabular}{ l c c c c}    
      & \multicolumn{2}{c}{\textbf{Link Prediction}} & \multicolumn{2}{c}{ \textbf{Node Classification}} \\ \toprule
       & {\textbf{GDPR}} & {\textbf{PGR}} & {\textbf{Cora}} & {\textbf{Ogbn-Arxiv}}\\
     
      \textbf{Model}    & \textbf{F1}  & \textbf{F1}  & \textbf{Acc}     & \textbf{Acc} \\ \midrule
   \textbf{GTNN} & 82.4  & 93.4 & 91.5 & 71.6 \\ \midrule
   \textbf{CurGraph} & 81.0 & 80.3 & 88.6 & 68.7 \\
\textbf{SL}  & 84.1 & 94.5 & 90.4 & 71.8 \\
 \textbf{Trend-SL}  & 84.6 & 94.5 & 90.4 &  71.5  \\
  \textbf{MCCL } & \textbf{85.7*}  & \textbf{95.2*} & \textbf{98.2*} & \textbf{76.5*} \\ \bottomrule

    \end{tabular}%
    \caption{Performance of curriculum models on \textbf{GDPR} and \textbf{PGR} datasets for link prediction, and \textbf{Cora} and \textbf{Ogbn-Arxiv} for node classification. The base model for all curriculum learning approaches is GTNN, which has a high score of F1 and Accuracy on the datasets using standard training. MCCL performs best compared to the other curricula methods. Asterisk marks (*) indicate significantly better performance compared to all other competing models.}
    \label{tab:curricula}%

\end{table}%

\subsection{Main Results}

Table \ref{tab:curricula} shows the performance of the proposed MCCL method against other curriculum learning approaches. The results shows that applying curricula to the base model (GTNN) further improves its performance by 3.3 and 1.8 absolute points in F1 on GDPR and PGR datasets respectively, indicating the importance of curriculum learning for training GNNs. The corresponding improvement on Cora and Ogbn-Arxiv datasets are 6.7 and 4.9 absolute points in accuracy. 

%
In addition, MCCL outperforms other curriculum learning approaches. Furthermore, as MCCL increasingly introduces more training instances at each iteration, it shows an overall faster training time compared to the other curriculum learning models, which iterate through all training examples at every iteration. See Section~\ref{sec:time_comp} for detail analysis on time complexity of different models.


\begin{table*}\small
\centering
        \begin{tabular}{l l l  c }
              \multicolumn{4}{c}{\textbf{Link Prediction}} \\ \toprule
            \textbf{Model} & \textbf{Index} & \textbf{Transition}  & \textbf{Avg F1}\\
             & \textbf{Order} & \textbf{Order}  & \\
            \midrule
            \textbf{GTNN}  & --    & --  &  87.9\\ \midrule
            \multirow{4}{*}{\textbf{MCCL: Model-based}} & desc & max     & 89.4 \\
              & desc & min     & 89.3 \\
            & asc  & max    & \textbf{89.9}\\ 
             & asc  & min    & 88.7 \\\midrule
             \multirow{4}{*}{\textbf{MCCL: Index-based}} & desc & max    &  90.1\\
               & desc & min  & 89.2\\
              & asc  & max     &  89.3\\
              & asc & min     & \textbf{90.4} \\
            \bottomrule
            \end{tabular}%
\hfill
        \begin{tabular}{l l l  c }
          \multicolumn{4}{c}{\textbf{Node Classification}}  \\ \toprule
        \textbf{Model} & \textbf{Index} & \textbf{Transition}  & \textbf{Avg Acc}\\
         & \textbf{Order} & \textbf{Order}  & \\
        \midrule
        \textbf{GTNN}  & --    & --  &  81.6\\ \midrule
        \multirow{4}{*}{\textbf{MCCL: Model-based}} & desc & max     &\textbf{ 87.3 }\\
          & desc & min     & 86.7 \\
        & asc  & max     & 87.1\\ 
         & asc  & min    & 86.5 \\\midrule
         \multirow{4}{*}{\textbf{MCCL: Index-based}} & desc & max    &  \textbf{87.0}\\
           & desc & min  & 86.3\\
          & asc  & max     &  86.9\\
          & asc & min     & 86.7 \\
        \bottomrule
        \end{tabular}%
    \caption{Ablation analysis on the order by which training examples are sorted for complexity indices (\underline{asc}ending versus \underline{desc}ending in the Index Order column, see line 2 in Algorithm~\ref{algo:ccl_multi_view}), the training mechanism (model- versus index-based in the Model column, see line 7 in Algorithm~\ref{algo:ccl_multi_view}) and the type of learning transition (easy-to-hard (Min error) versus hard-to-easy (Max error) in the Transition Order column, see lines 9--13 in Algorithm~\ref{algo:ccl_multi_view}).}
    \label{tab:ab_study}
\end{table*}

\section{Multiview Curricula Introspection}
We perform several ablation studies on the MCCL model, investigating genuine complexity indices compared to random ordering, multiview curricula versus anti-curricula, the impact of complexity indices in training, and the model's time complexity.

\subsection{Model Prioritizes Genuine Complexity Indices over Random Ordering}
In curriculum learning, effective training depends on the scheduler, which determines the set of examples and their order for training at each iteration. Hence, the performance of the model largely depends on the scheduling criteria used for training. To determine if our model can indeed prioritize better indices, we added a {\em fake} index named ``Random'' index to the list of our complexity indices. Training examples were randomly ordered in the Random index. We re-ran our model and checked whether it selects the Random index for training at any iteration. On Cora and Ogbn-Arxiv datasets, model selects the Random index at 17.6\% and 12.8\% of its training iterations. On PGR, the model never selects the Random index, and on GDPR, model selects the Random index at 8\% of its training iterations.  Specifically, toward the end of training at iterations [39, 46, 58, 71, 76, 83, 91, 93] with the best F1-score of 85.5\% obtained at iteration 76. The fact that the model do not often select the Random index at many iterations is effectively inline with the core principle of curriculum learning--learning materials should be gradually learned in a properly-planned order. This sanity check indicates that the model prioritizes genuine complexity indices over random ordering. 


\subsection{Multiview Curricula vs. Anti-Curricula}
We study the effect of different criteria in MCCL framework through ablation analysis on 
(a): the order by which training examples are sorted with respect to their complexity scores for each index (descending versus ascending, see line 2 in Algorithm~\ref{algo:ccl_multi_view}), 
(b): the mechanism by which our framework prioritizes indices (model-based versus index-based, see line 7 in Algorithm~\ref{algo:ccl_multi_view} and Section~\ref{sec:prioritizing}), and 
(c): the type of learning transition in our framework (easy-to-hard versus hard-to-easy transition, see lines 9--13 in Algorithm~\ref{algo:ccl_multi_view}).

Table~\ref{tab:ab_study} shows the result of this ablation analysis averaged over the PGR and GDPR datasets for link prediction, and Cora and Ogbn-Arxiv datasets for node classification respectively. The corresponding results for each dataset is reported in Appendix~\ref{sec:appendix_abl}. Overall, the ascending order results in the best average F1 score for link prediction while descending order performs better for node classification.
 In addition, in model-based training, hard-to-easy (max) transition order is more effective than easy-to-hard (min) transition order across both tasks. This is perhaps because harder examples are superior at helping the model find better local minima at the early stages of training. We also observe that easy-to-hard (min) transition for index-based training results in higher average F1-score than hard-to-easy (max) transition of the model-based training. This is because, in case of index-based ordering, the difficulty scores (which are obtained from indices) may provide a more accurate estimation of easiness to the model than hardness, i.e. easy examples are likely easy for the model in both ordering but this may not be true for hard examples.

\subsection{Index Contributions to Training}
To study the contributions of different complexity indices in the training process, we divide training iterations into three phases and create the histograms that show the number of times that each index is chosen at different stages of training: (i) Initial, (ii) Middle, and (iii) End phases of the training. 


\begin{figure*}
    \centering
    \subfigure[GDPR dataset]
    {\includegraphics[scale=0.49]{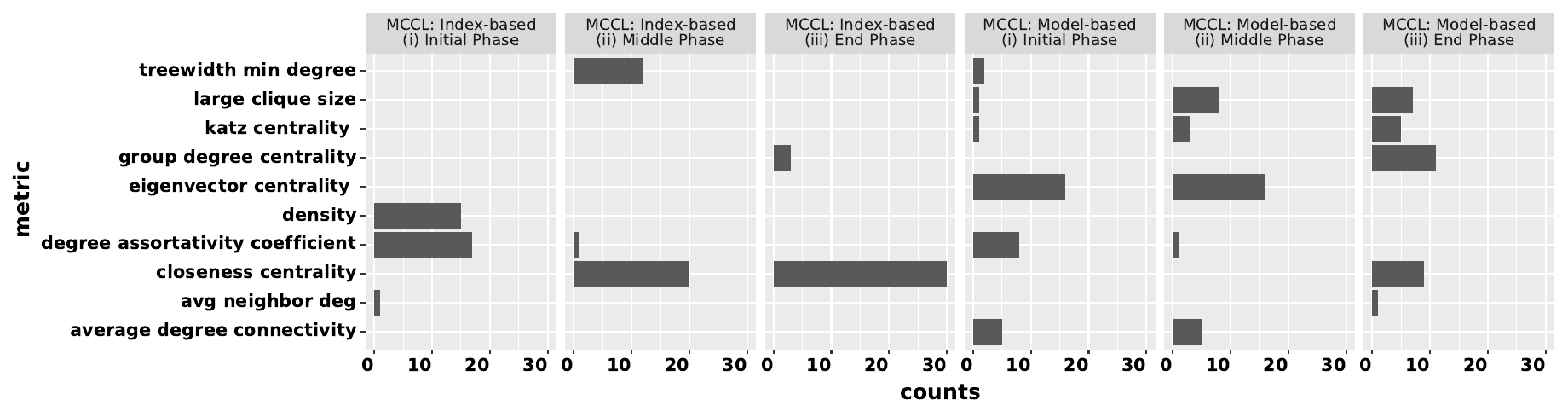}\label{ablation_gdpr_metric}}
    
    \subfigure[PGR dataset]
    {\includegraphics[scale=0.49]{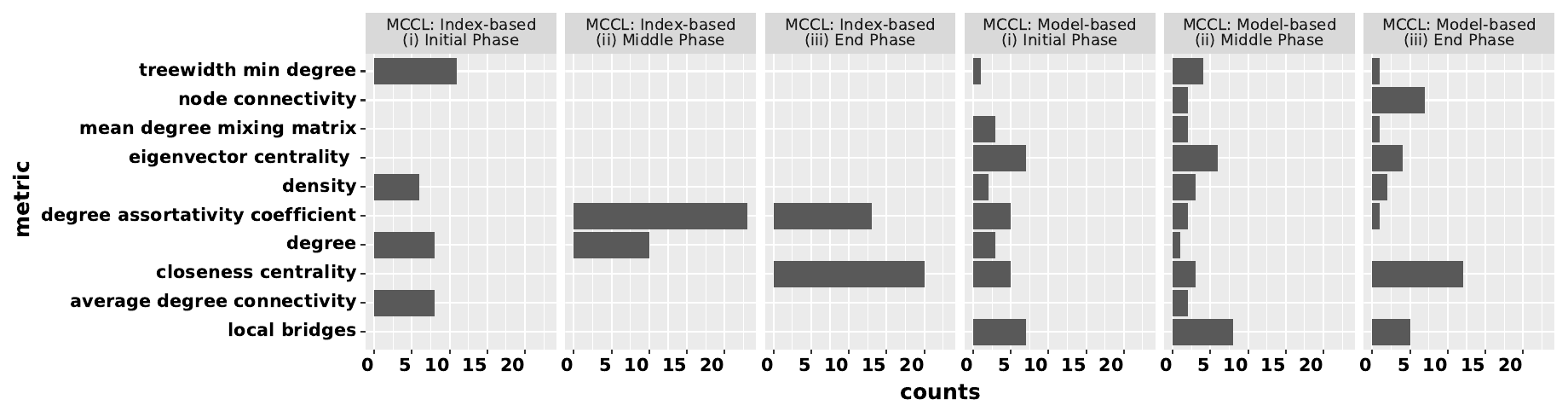}\label{ablation_pgr_metric}}
    
    \subfigure[Cora dataset]
    {\includegraphics[scale=0.49]{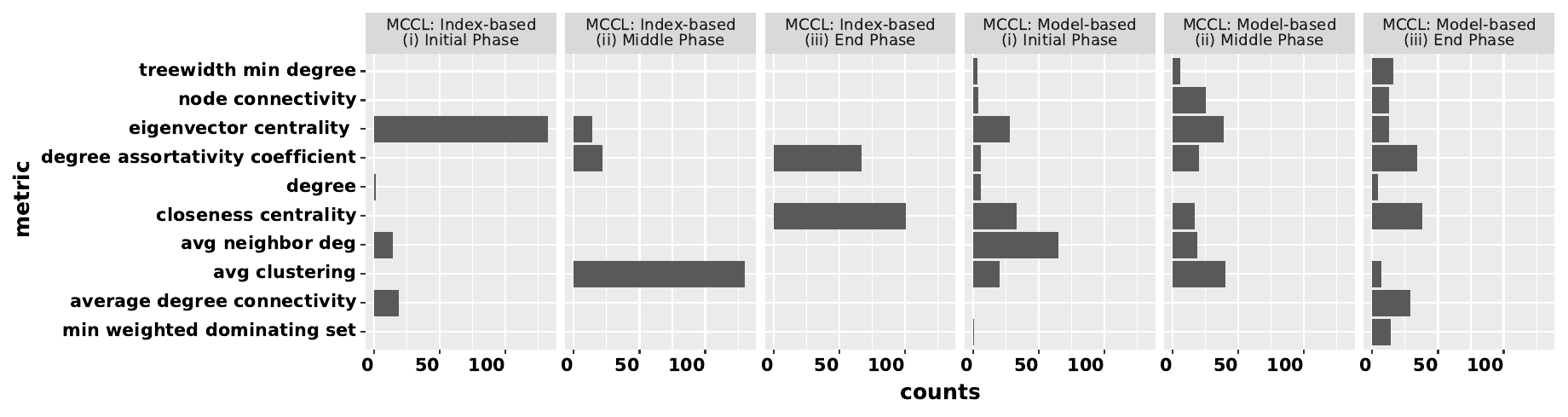}\label{ablation_cora_metric}}
    
    \subfigure[Ogbn-Arxiv dataset]
    {\includegraphics[scale=0.49]{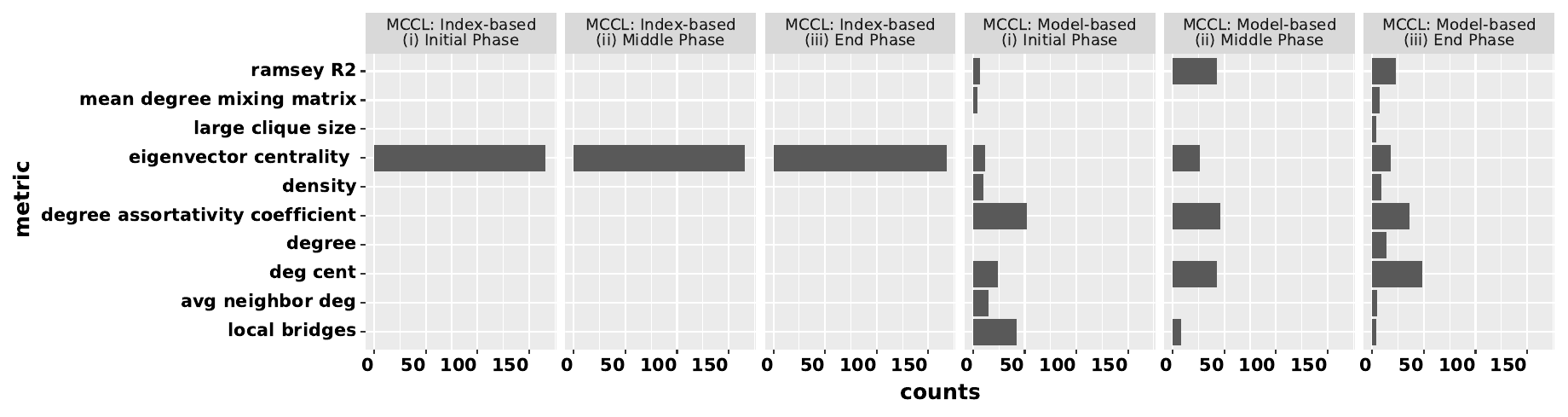}\label{ablation_arxiv_metric}}
    
    \caption{\small{Histogram of the different indices chosen by the best-performing MCCL model for both index-based (where samples are selected for training merely based on their difficulty scores obtained from indices) and model-based (where samples are selected based on instantaneous loss of the model) approaches across three datasets at the beginning, middle and end of training; see best models in Table~\ref{tab:gdpr_ab_study}, Appendix~\ref{sec:appendix_abl}. 
    Dominance of a metric depends on the internal structure of the graph, underlying task, stage of training, and model behavior during training.
    (a): on GDPR, model mainly focuses on density- and degree-based indices (which operate locally around nodes) at the beginning and centrality-based indices (which operate globally at graph level) at later stages of index-based training (left three plots). In case of model-based training (right three plots), both degree- and centrality- based indices were selected throughout the training. 
    (b): On PGR, we observed the same pattern except that locality indices such as number of local bridges were prioritized in case of model-based training.    
    (c): On Cora, the model merely focuses on the degree-based metrics throughout its index-based training, and focuses on a mix of different indices at early stages of its model-based training and closeness centrality at the end of training.
    (d): On Ogbn-Arxiv, the model only focuses on eigenvector centrality metric throughout its index-based training and focuses on the local bridge and degree assortativity indices in the initial phase of the model-based training. In the middle and final phases, model-based training mainly focuses on degree based indices.} }
    \label{fig:ablation_metric}
\end{figure*} 

Figures~\ref{fig:ablation_metric} shows the results for different indices chosen by the best-performing MCCL model\footnote{According to the results in Tables~\ref{tab:gdpr_ab_study}~and~\ref{tab:cora_ab_study} in Appendix~\ref{sec:appendix_abl}.} for both index-based (where the criterion for selecting samples is merely based on their difficulty scores obtained from indices) and model-based (where the criterion for selecting samples is based on instantaneous loss) approaches across our four datasets.
Our key observation was that MCCL mainly focused on indices that operate locally around nodes (such as 
density- or degree-based indices) at early stages of training and then focused on indices that operate globally at graph level (such as centrality-based indices) at later stages of training for both index-based and model-based training mechanisms. Additionally, we observed greater diversity in the sets of prioritized indices in case of the model-based training mechanism, which indicates MCCL encourages learning from diverse views of difficulty during training. This is mainly because the model-based training mechanism in MCCL allows the GNN to directly contribute in scheduling indices through its loss dynamics during training.  

In addition, we note that on the Cora dataset the model merely focused on degree-based metrics throughout its training in case of index-based training and, used a smaller set of fundamentally-different indices in case of model-based training. On the Ogbn-Arxiv dataset, the model focuses only on the eigenvector centrality index throughout its training in case of index-based training and focuses on connectivity and centrality indices in the model-based training. Further analysis on this model behavior is the subject of our future work.  

Overall, density, degree indices including degree and degree assortativity coefficient, and centrality indices including closeness centrality, group degree centrality and eigenvector centrality indices are often prioritized by the model across different mechanisms and datasets.

\subsection{MCCL Has the Lowest Time Complexity}\label{sec:time_comp}
Let $n$ be the number of training examples and $e$ be the maximum number of iterations for training a neural network. The total of number of forward and backward passes required to train GTNN, SL, Trend-SL is $2\times n\times e$. In contrast, MCCL trains from only a fraction of training examples at each iteration and do not need to ``see'' all the training examples at each iteration. If model-based mechanism is used to prioritize important indices for training, the total number of forward and backward passes of MCCL is $3 \times \sum_i (n \times \frac{i}{e})$, which amounts to $1.5\times n\times(e-1)$; note that model-based approach requires $\sum_i (n\times \frac{i}{e}) = \frac{n\times(e-1)}{2}$ additional forward passes to select the best index. In case of the index-based training mechanism, no additional forward pass is required, resulting in a total of $2\times \sum_i (n\times \frac{i}{e})$ passes, which amounts to $n\times(e-1)$ passes. In either case, MCCL has lower time complexity than  other baselines.
The turnaround time of our model ranges from 10 minutes to 2.5 hours, depending on the size of the input dataset.

\section{Related Work}

Curriculum learning \cite{bengio2009curriculum} aims to improve the generalizability of a model by gradually training it with easy examples followed by hard ones. 
\citet{castells2020superloss} introduced a generic loss function called SuperLoss (SL) which can be added on top of any target-task loss function to dynamically weight the training samples according to their difficulty for the model using a batch-wise threshold. 
\citet{zhou2020curriculum} proposed dynamic instance hardness to determine the difficulty of an instance with running average of the hardness metric over training history.

Curriculum learning has been investigated in NLP~\citep{elman1993learning,sachan-xing-2016-easy,settles2016trainable,amiri-etal-2017-repeat,platanios2019competence,amiri2019neural,zhang-etal-2019-curriculum,lalor-yu-2020-dynamic,xu-etal-2020-curriculum,chu2021cuco,liu-etal-2021-competence,kreutzer-etal-2021-bandits-dont,agrawal2022imitation,maharana-bansal-2022-curriculum}. 
Specifically, \citet{settles2016trainable,amiri-etal-2017-repeat} proposed spaced repetition-based curricula based on psycholinguistic theory where the training data is scheduled by increasing intervals of time between consecutive reviews of previously learned data samples. \citet{zhang-etal-2019-curriculum} investigated curriculum learning for domain adaptation in neural machine translation, where samples were grouped and ranked based on their similarity score such that more similar samples are seen earlier and more frequently during training.  
\citet{platanios2019competence} proposed an approach to use competency function using rarity of words or length of a sentence for neural machine translation and inspired \citet{liu-etal-2021-competence} to define a curriculum based on multi-modal (text and image) data to choose which modality should be used for training. The model uses sample perplexity at batch level to select the modality for training. Linguistic features such as word rarity or length of sentence in~\cite{platanios2019competence} and sample perplexity in~\citep{liu-etal-2021-competence} were used as measures of difficulty. 
\citet{xu-etal-2020-curriculum} designed a curriculum learning approach for NLP tasks using cross-view of training data to identify easy and hard examples and rearrange the examples during training. 
Other works of curriculum learning in NLP focused on machine translation and language understanding. \citet{agrawal2022imitation} developed a framework to train non-autoregressive sequence-to-sequence model to edit text where a curriculum is designed to first perform easy-to-learn edits followed by increasing difficulty of training samples. 
\citet{maharana-bansal-2022-curriculum} designed several curriculum learning approaches using teacher-student model where the teacher model calculates the difficulty of each training example using question-answering probability, variability, and out-of-distribution measures.

Curriculum learning for graph data is an emerging area of research.
\citet{chu2021cuco}  explored curriculum learning approach in the self-supervision settings where the difficulty measure evaluates the difficulty of negative samples which is calculated based in the embeddings's similarity between positive and negative examples. 
\citet{wang2021curgraph} proposed a curriculum based subgraph classification approach, CurGraph, which first obtains graph-level embeddings via unsupervised GNN method and then uses neural density estimator to model embedding distributions. The difficulty scores of graphs are calculated by a predefined difficulty measure based on the inter- and intra-class distribution of sub-graph embeddings.
In~\citet{nidhi-etal-2022-gtnn}, we extended the SuperLoss approach developed in~\citep{castells2020superloss} by introducing a curriculum learning framework that dynamically adjusts the difficulty of samples during training with respect to the loss trajectory. We demonstrated the effectiveness of incorporating this strategy in graph curriculum learning settings. 
Previous work in graph curriculum learning has employed a single criterion of difficulty in their curriculum learning framework. Our work uses multiple criteria for curriculum learning on graphs. We encourage readers to see \citep{2023arXiv230202926L} for a survey on graphs curriculum learning approaches.

Finally, in terms of datasets, \citet{sousa2019silver} developed the PGR dataset, which we used in our experiments. They developed a transformer model to identify the relation between biomedical entities, genes and phenotypes, from scientific PubMed articles. For relation classification authors considered a pair of entities and the context from the corresponding sentence in which both entities occur.

\section{Conclusion and Future work}
We present a novel curriculum learning approach for training graph neural networks. Our approach combines well-established graph complexity indices (views) obtained from graph theory and demonstrates the effectiveness of learning from diverse difficulty views for the tasks of link prediction and node classification. Our approach improves over the state-of-the-art techniques for curriculum learning on graphs across several datasets. Ablation studies show that the model prioritizes genuine complexity indices over to random ordering, and effectively uses and learn multiview complexity indices in both curricula and anti-curricula settings, and has lower time complexity that competing models.
In future, we will extend our approach to other graph processing tasks, focusing on NLP applications such as clustering and community detection, and investigate the effect of graph complexity indices in such tasks.    

\section*{Limitation}
Calculating complexity indices for large-scale graphs can be computationally expensive and time consuming. Some of the complexity indices show longer turnaround time when computed for denser areas in the graphs. In addition, as we mentioned in the paper, although we made sure our framework and implementation allows adding any number of additional indices in a modular way, there might be other effective complexity indices that are not included in this investigation. Furthermore, it should be noted that the model has been exclusively tested on graphs where nodes contain textual content, which may limit its application to more general graph types. Finally, the model has not been applied to other graph-based tasks such as clustering and graph-level classification. 
\newpage
\bibliography{anthology,custom}
\newpage
\clearpage
\appendix
\section{Graph Difficulty Indices}\label{sec:appendix}

    Below are the list of 26 indices which we consider for  Multiview Competence-based Curriculum Learning (MCCL) approach. All these indices are computed on the subgraph of the node or an edge. These definition and code to calculate the indices, we used Networkx package \cite{hagberg2008exploring}.

\begin{itemize}
 
    \item \textbf{* Degree:} The number of immediate neighbors of a node in a graph. 
    
    \item \textbf{* Treewidth min degree:} The treewidth of an graph is an integer number which quantifies,  how far the given graph is from being a tree. 
    
    \item \textbf{* Average neighbor degree:} Average degree of the neighbors of a node is computed as: 
    \begin{eqnarray*}
    \frac{1}{|\mathcal{N}_i|}\sum_{j\in \mathcal{N}_i} k_j
    \end{eqnarray*}
    where $\mathcal{N}_i$ is the set of neighbors of node $i$ and $k_j$ is the degree of node $j$. 
    
    \item \textbf{* Degree mixing matrix:} Given the graph, it calculates joint probability, of occurrence of node degree pairs. Taking the mean, gives the degree mixing value representing the given graph.
    
    \item \textbf{* Average degree connectivity:} Given the graph, it calculates the average of the nearest neighbor degree of nodes with degree $k$. We choose the highest value of $k$ obtained from the calculation and used its connectivity value as the complexity index score.

\end{itemize}

     \begin{table}[h]
\footnotesize
  \centering
    \begin{tabular}{p{3.78cm}|l}
    \hline
    \textbf{Degree based} & \textbf{Computing based} \\
    degree & ramsey R2 \\
    treewidth min degree  & average clustering   \\
    degree mixing matrix & resource allocation index \\ 
    average neighbor degree             & \textbf{Basic properties} \\
    average degree connectivity         & density    \\
    degree assortativity coefficient     & local bridges  \\
    \textbf{Centrality}                & number of nodes \\
     katz centrality                    & number of edges \\
    degree centrality                   & large clique size \\
    closeness centrality                &  common neighbors \\
    eigenvector centrality              & \textbf{Connectivity}\\
     group degree centrality             & subgraph connectivity \\ 
     \textbf{Flow property}             &  local node connectivity\\
    min weighted dominating set         &  \\
    min weighted vertex cover           &  \\ 
    min edge dominating set \\
    min maximal matching \\
    \hline
    \end{tabular}%
      \caption{Complexity indices used to capture diverse characteristics of a graph structure. For different graph structure, some index may represent the same order with another metric when sorted. To avoid this redundancy, we rank training data for each index and find Pearson corelation between each such ranking. We then used k-means to find the clusters. }

  \label{tab:metric_info}%
\end{table}%

\begin{itemize}
    \item \textbf{* Degree assortativity coefficient:} Given the graph, assortativity measures the similarity of connections in the graph with respect to the node degree. 
   
   \item \textbf{* Katz centrality:} The centrality of a node, $i$, computed based on the centrality of its neighbors $j$. Katz centrality computes the relative influence of a node within a network by measuring taking into account the number of immediate neighbors and
    number of walks between node pairs. It is computed as follows:
    \begin{eqnarray*}
    x_{i} = \alpha\sum_{j}A_{ij}x_{j}+\beta
    \end{eqnarray*}
    where $x_i$ is the Katz centrality of node $i$, $A$ is the adjacency matrix of Graph $G$ with eigenvalues $\lambda$. The parameter $\beta$ controls the initial centrality and $\alpha$ $<$ 1 / $\lambda_{max}$.
    
    \item \textbf{Degree centrality:} Given the graph, the degree centrality for a node is the fraction of nodes connected to it. 
    
    \item \textbf{* Closeness centrality:} The closeness of a node is the distance to all other nodes in the graph or in the case that the graph is not connected to all other nodes in the connected component containing that node. Given the subgraph and the nodes, added the values of the nodes to find the complexity index value. 
    
\textbf{\item \textbf{* Eigenvector centrality:}} Eigenvector centrality computes the centrality for a node based on the centrality of its neighbors. The eigenvector centrality for node i is $Ax$ = $\lambda x $. where $A$ is the adjacency matrix of the graph $G$ with eigenvalue $\lambda$.
    
    \item \textbf{* Group Degree centrality:} Group degree centrality of a group of nodes S is the fraction of non-group members connected to group members. 
    
    \item \textbf{Ramsey R2:} This computes the largest clique and largest independent set in the graph $G$. We calculate the index value by multiplying number of largest cliques to number of largest independent set. 
    
    \item \textbf{* Average clustering:} The local clustering of each node in the graph $G$ is the fraction of triangles that exist over all possible triangles in its neighborhood. The average clustering coefficient of a graph $G$ is the mean of local clusterings.
    
    \item \textbf{Resource allocation index:} For nodes $i$ and $j$ in a subgraph, the resource allocation index is defined as follows: 
    \begin{eqnarray*}
    \sum_{k \in (\mathcal{N}_i\bigcap\mathcal{N}_j)}\frac{1}{|\mathcal{N}_k|},
    \end{eqnarray*}
    which quantifies the closeness of target nodes based on their shared neighbors.
    
    \item \textbf{* Subgraph density:} The density of an undirected subgraph is computed as follows:
    \begin{eqnarray*}
    \frac{e}{v(v-1)},
    \end{eqnarray*}
    where $e$ is the number of edges and $v$ is the number of nodes in the subgraph.

    \item \textbf{* Local bridge:} A local bridge is an edge that is not part of a triangle in the subgraph. We take the number of local bridges in a subgraph as a complexity score.
    
    \item \textbf{Number of nodes:} Given the graph $G$, number of nodes in the graph is chosen as the complexity score.
    
    \item \textbf{Number of Edges:} Given the graph $G$, number of edges in the graph is chosen as the complexity score.
    
    \item \textbf{* Large clique size:} Given the graph $G$, the size of a large clique in the graph is chosen as the complexity score.
    
    \item \textbf{Common neighbors:} Given the graph and the nodes, it finds the number of common neighbors between the pair of nodes. We chose number of common neighbors as the complexity score. 
    
    \item \textbf{* Subgraph connectivity:} is measured by the {\em minimum} number of nodes that must be removed to disconnect the subgraph.
    
    \item \textbf{Local node connectivity:} Local node connectivity for two non adjacent nodes s and t is the minimum number of nodes that must be removed (along with their incident edges) to disconnect them. Given the subgraph and the nodes, gives the single value which we used as complexity score.

    \item \textbf{Minimum Weighted Dominating Set}: For a graph $G = (V,E)$, the weighted dominating set problem is to find a vertex set $\mathcal{S} \subseteq V$ such that when  each vertex is associated with a positive number, the goal is to find a dominating set with the minimum weight.
    
    \item \textbf{Weighted vertex cover index:} The weighted vertex cover problem is to find a vertex cover $\mathcal{S}$--a set of vertices that include at least one endpoint of every edge of the subgraph--that has the minimum weight. This index and the weight of the cover $\mathcal{S}$ is defined by $\sum_{s\in\mathcal{S}} w(s)$, where $w(s)$ indicates the weight of $s$. Since $w(s)=1, \forall s$ in our unweighted subgraphs, the problem will reduce to finding a vertex cover with minimum cardinality. 
    
    \item \textbf{Minimum edge dominating set:} Minimum edge dominating set approximate solution to the edge dominating set.
    \item \textbf{Minimum maximal matching:} Given a graph G = (V,E), a matching M in G is a set of pairwise non-adjacent edges; that is, no two edges share a common vertex. That is, out of all maximal matchings of the graph G, the smallest is returned. We took the length of the set as the complexity index.

\end{itemize}

\section{Multiview Curricula Ablation Analysis}\label{sec:appendix_abl}

\begin{table*}[t]
  \centering
    \begin{tabular}{l l l c c c c }
    \textbf{Dataset}  &  \textbf{Model} & \textbf{Index Order} & \textbf{Transition Order}  & \textbf{P} & \textbf{R} & \textbf{F1} \\
    \hline
    PGR   & GTNN  & --    & --    & 93.6  & 93.2  & 93.4 \\\hline
    PGR   & MCCL: Model-based & descending & max   & 95.8 & 93.2  & 94.5 \\
    PGR   & MCCL: Model-based & descending & min   & 93.3  & 94.6  & 94.0 \\
    PGR   & MCCL: Model-based & ascending  & min   & 95.9  & 94.6 & \textbf{95.2} \\
    PGR   & MCCL: Model-based & ascending & max   & 97.2  & 93.2  & \textbf{95.2} \\
    
    PGR   & MCCL: Index-based  & descending & min   & 94.5 & 93.2  & 93.9 \\
    PGR   & MCCL: Index-based  & ascending & max   & 97.2 & 93.2  & \textbf{95.2} \\
    PGR   & MCCL: Index-based  & ascending & min   & 95.9 & 94.6  & \textbf{95.2} \\
    PGR   & MCCL: Index-based  & descending & max   & 97.1 & 91.9  & 94.4 \\
    \hline\hline
    GDPR  & GTNN  & --    & --    & 77.1  & 88.5  & 82.4 \\\hline
    GDPR  & MCCL: Model-based & descending & max & 80.6 & 88.2 & 84.3 \\
    GDPR  & MCCL: Model-based & descending & min   & 83.0  & 86.3  & 84.6 \\
    GDPR  & MCCL: Model-based & ascending  & min   & 78.9  & 85.7  & 82.1 \\
    GDPR  & MCCL: Model-based & ascending & max   & 80.6  & 89.0  & 84.6 \\
    GDPR  & MCCL: Index-based  & descending & min & 82.4 & 86.5 & 84.4 \\
    GDPR  & MCCL: Index-based & descending & max   & 86.2  & 85.2  & \textbf{85.7} \\
    GDPR  & MCCL: Index-based  & ascending & min & 84.6 & 86.5 & 85.5 \\
    GDPR  & MCCL: Index-based & ascending & max   & 84.2  & 82.6  & 83.4 \\
    \hline
    \end{tabular}%
      \caption{Ablation analysis on PGR and GDPR datasets with respect to the order by which training examples are sorted for complexity indices (ascending versus descending, see Index Order column and line 2 in Algorithm~\ref{algo:ccl_multi_view}), the mechanism by which indices are prioritized (model-based versus index-based, see Model column and line 7 in Algorithm~\ref{algo:ccl_multi_view}) and the type of learning transition (easy-to-hard (Min error) versus hard-to-easy (Max error) transition, see Transition Order column and lines 9--13 in Algorithm~\ref{algo:ccl_multi_view}).}

  \label{tab:gdpr_ab_study}%
\end{table*}%

\begin{table*}[t]
  \centering
    \begin{tabular}{l l l c c c c }
    \textbf{Dataset}  &  \textbf{Model} & \textbf{Index Order} & \textbf{Transition Order}   & \textbf{Acc} \\
    \hline
    Ogbn-Arxiv   & GTNN  & --    & --    & 71.6 \\\hline
    Ogbn-Arxiv   & MCCL: Model-based & descending & max   &  \textbf{76.5}\\
    Ogbn-Arxiv   & MCCL: Model-based & descending & min   & 76.0 \\
    Ogbn-Arxiv   & MCCL: Model-based & ascending  & max   &  76.4 \\
    Ogbn-Arxiv   & MCCL: Model-based & ascending & min   & 76.2  \\
    
    Ogbn-Arxiv   & MCCL: Index-based  & descending & max   & 76.2 \\
    Ogbn-Arxiv   & MCCL: Index-based  & descending & min   &  76.3\\
    Ogbn-Arxiv   & MCCL: Index-based  & ascending & max   & 76.4 \\
    Ogbn-Arxiv   & MCCL: Index-based  & ascending & min   &  76.0\\
    \hline\hline
    Cora  & GTNN  & --    & --    & 91.5\\\hline
    Cora  & MCCL: Model-based & descending & max & \textbf{98.2}\\
    Cora  & MCCL: Model-based & descending & min   &  97.4 \\
    Cora  & MCCL: Model-based & ascending  & max   &  97.8\\
    Cora  & MCCL: Model-based & ascending & min   &  96.7\\
    Cora  & MCCL: Index-based  & descending & max &  97.8\\
    Cora  & MCCL: Index-based & descending & min   & 96.3\\
    Cora  & MCCL: Index-based  & ascending & max & 97.4 \\
    Cora  & MCCL: Index-based & ascending & min   &  97.4\\
    \hline
    \end{tabular}%
      \caption{Ablation analysis on Ogbn-Arxiv  and Cora datasets with respect to the order by which training examples are sorted for complexity indices (ascending versus descending, see Index Order column and line 2 in Algorithm~\ref{algo:ccl_multi_view}), the mechanism by which indices are prioritized (model-based versus index-based, see Model column and line 7 in Algorithm~\ref{algo:ccl_multi_view}) and the type of learning transition (easy-to-hard (Min error) versus hard-to-easy (Max error) transition, see Transition Order column and lines 9--13 in Algorithm~\ref{algo:ccl_multi_view}).}

  \label{tab:cora_ab_study}%
\end{table*}%

The results of our study the effect of different curriculum criteria in our competence-based multiview curriculum learning framework. We conduct ablation analysis on 
(a): the order by which training examples are sorted with respect to their complexity scores for each index (descending versus ascending, see line 2 in Algorithm~\ref{algo:ccl_multi_view}), 
(b): the mechanism by which our framework prioritizes indices (model-based versus index-based, see line 7 in Algorithm~\ref{algo:ccl_multi_view} and Section~\ref{sec:prioritizing}), and 
(c): the type of learning transition in our framework (easy-to-hard versus hard-to-easy transition, see lines 9--13 in Algorithm~\ref{algo:ccl_multi_view}).

\end{document}